\documentclass[letterpaper, 10pt, conference]{ieeeconf}

\IEEEoverridecommandlockouts 

\usepackage{graphics} 
\usepackage{epsfig} 
\usepackage{amsmath} 
\usepackage{amssymb}  
\usepackage{algorithm2e}
\usepackage{siunitx} 
\usepackage{pdfcomment}
\usepackage{xcolor}
\usepackage{subfigure}
\usepackage{tabularx}

\title{\LARGE \bf
Fast Simulation of Vehicles with Non-deformable Tracks
}

\author{Martin Pecka$^{1,2}$, Karel Zimmermann$^{2}$,  and Tom{\'a}{\v s} Svoboda$^{1,2}$
\thanks{$^{1}$Czech Technical University in Prague, Czech Institute of Informatics Robotics and Cybernetics}%
\thanks{$^{2}$Czech Technical University in Prague, Faculty of Electrical Engineering, Department of Cybernetics}%
}

\begin{document}

\maketitle
\thispagestyle{empty}
\pagestyle{empty}

\begin{abstract}
This paper presents a~novel technique that allows for both computationally fast and sufficiently plausible simulation of vehicles with non-deformable tracks.
The method is based on an effect we have called \textit{Contact Surface Motion}.
A~comparison with several other methods for simulation of tracked vehicle dynamics is presented with the aim to evaluate methods that are available off-the-shelf or with minimum effort in general-purpose robotics simulators.
The proposed method is implemented as a~plugin for the open-source physics-based simulator Gazebo using the Open Dynamics Engine.
\end{abstract}

\section{INTRODUCTION}

Tracked vehicles are often preferred over the wheeled ones in tasks where traversing complicated terrain is needed, such as in Urban Search and Rescue missions.
Tracks provide higher stability, better traction and help the vehicle traverse holes in the underlying terrain.

It is common in robotics research that the initial development of algorithms is first conducted in a~simulator or game engine to avoid excessive wear of the real vehicle.
In this phase, approximate simulation methods usually suffice, differing by the level of approximation and computation time.
General-purpose simulators like Gazebo, V-REP, Webots, MORSE and Actin are often used for this task~\cite{Drumwright2010}, providing various approximate motion models implemented in their physics engines (ODE, Bullet, Havoc).

Simulation of wheels is straightforward in these simulators, thus all of them provide means to simulate wheeled vehicles, including skid-steer motion of multi-wheel vehicles.
However, there is no straightforward approach for tracked vehicle simulation, thus this motion model is not available in most simulators.
After an exhaustive search, only two simulators were found providing a~tracked robot in its robot model library -- the commercial simulators Webots and V-REP.
However, none of these implementations is both plausible on difficult terrain and computationally light.

The most plausible and general simulation methods for rubber belts are based on finite elements analysis, where the belt is subdivided in many small elements that interact in a~defined way.
We omit this class of methods in this work due to their inherent excessive computational complexity which makes them impractical for quick algorithm prototyping.
Further argument for omitting these methods is that none of the most used open-source dynamics engines used in robotics supports simulation based on finite elements.

In this paper, we present a~novel technique for non-deformable tracks simulation, which we implemented in the open-source simulator Gazebo~\cite{gazebo-pr}.
The method provides a~fast, simple and plausible simulation of non-deformable tracks with minimal changes to the simulator code and no changes to its physics engine (ODE).
The method was successfully used in our prior work~\cite{Pecka-IROS-2016} for assessing safety of actions a~tracked vehicle can perform.
We would like to emphasize that our motivation is to have a fast and plausible method that can be easily integrated into existing robotics simulators and does not require implementation of state-of-the art physics engine components (which are usually absent in the robotics simulators).

We compare this method to other already known motion models.
Finally, we propose a~set of metrics that allow to compare the methods in terms of plausibility, computational time, and the range of track types that can be simulated by each of the respective techniques.

\begin{figure}[b]
    \centering
    \def\sz{0.29\columnwidth}
    \def\szz{0.322\columnwidth}
    
   	\includegraphics[height=\sz]{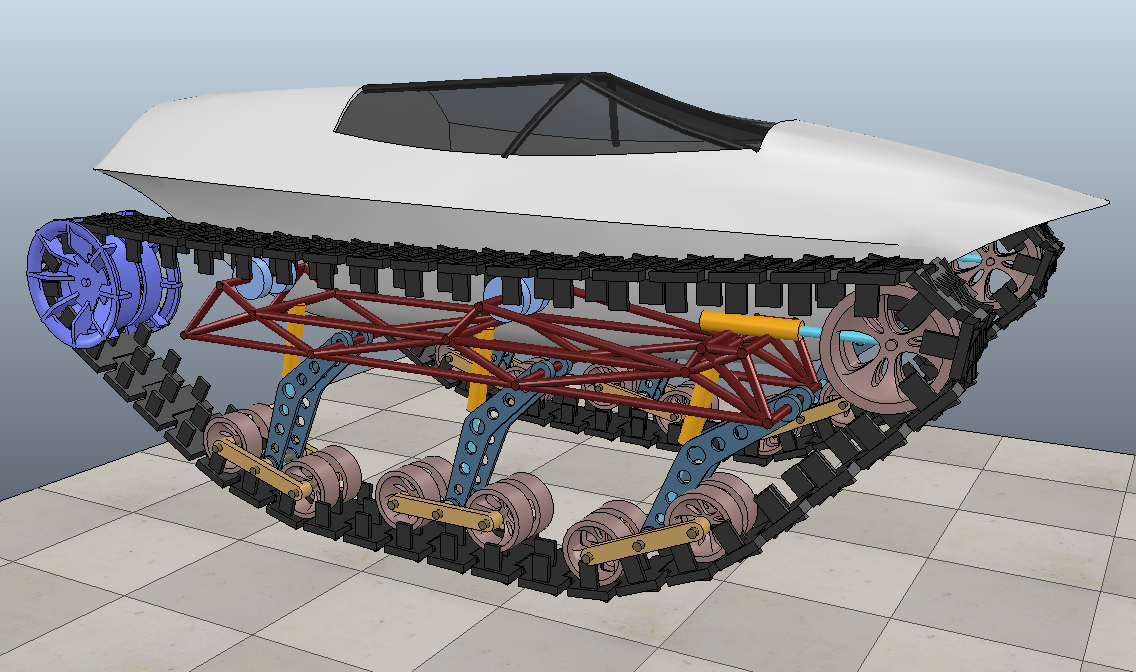}
   	\includegraphics[height=\sz]{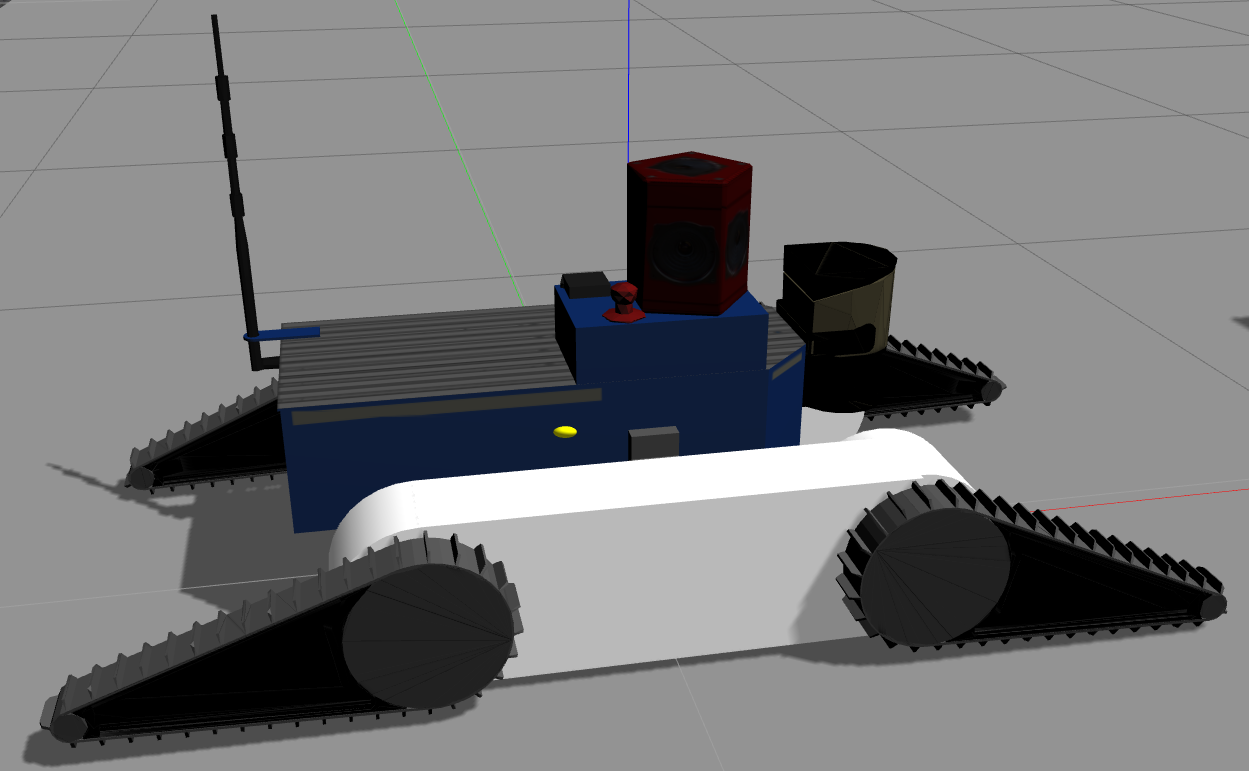} \\
   	\vspace{0.1cm}
   	\includegraphics[height=\szz]{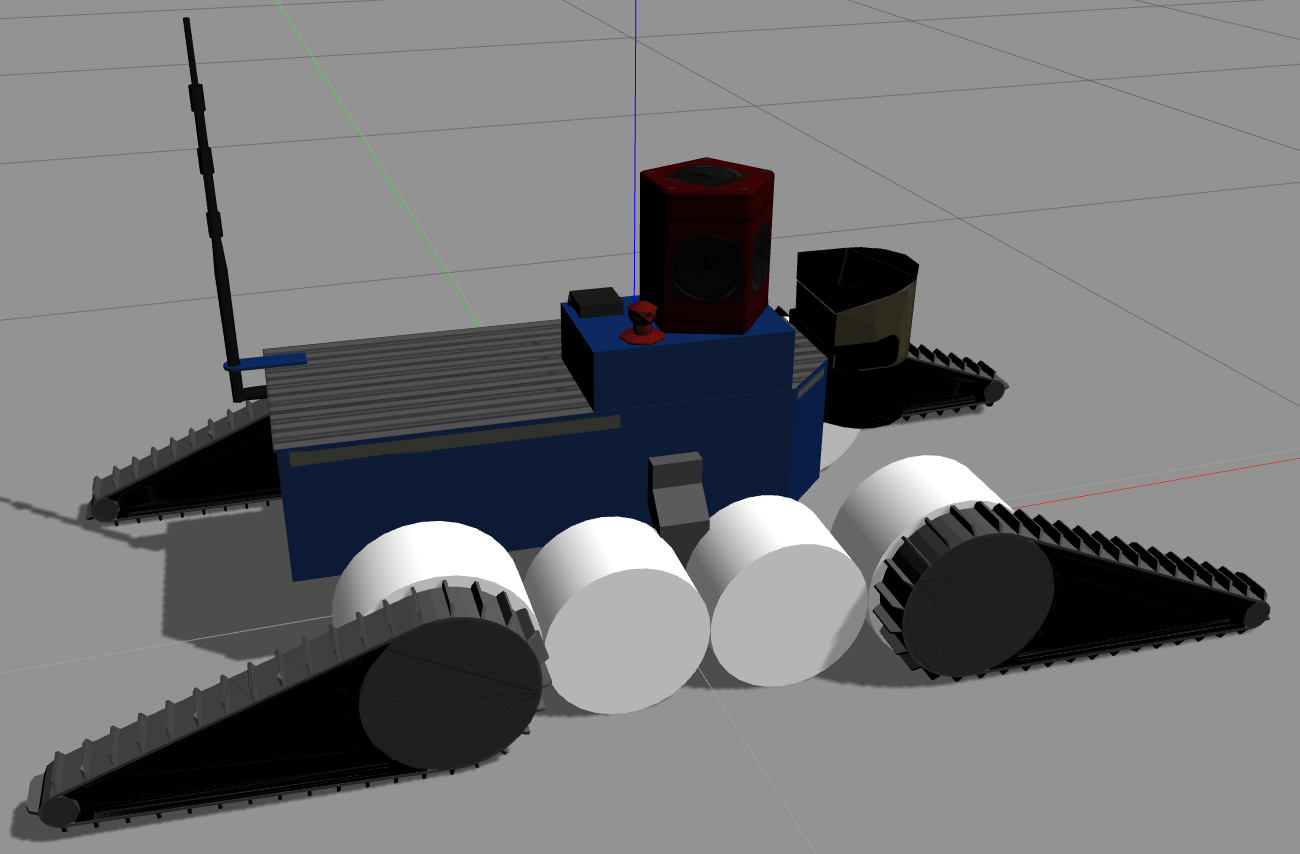}
   	\includegraphics[height=\szz]{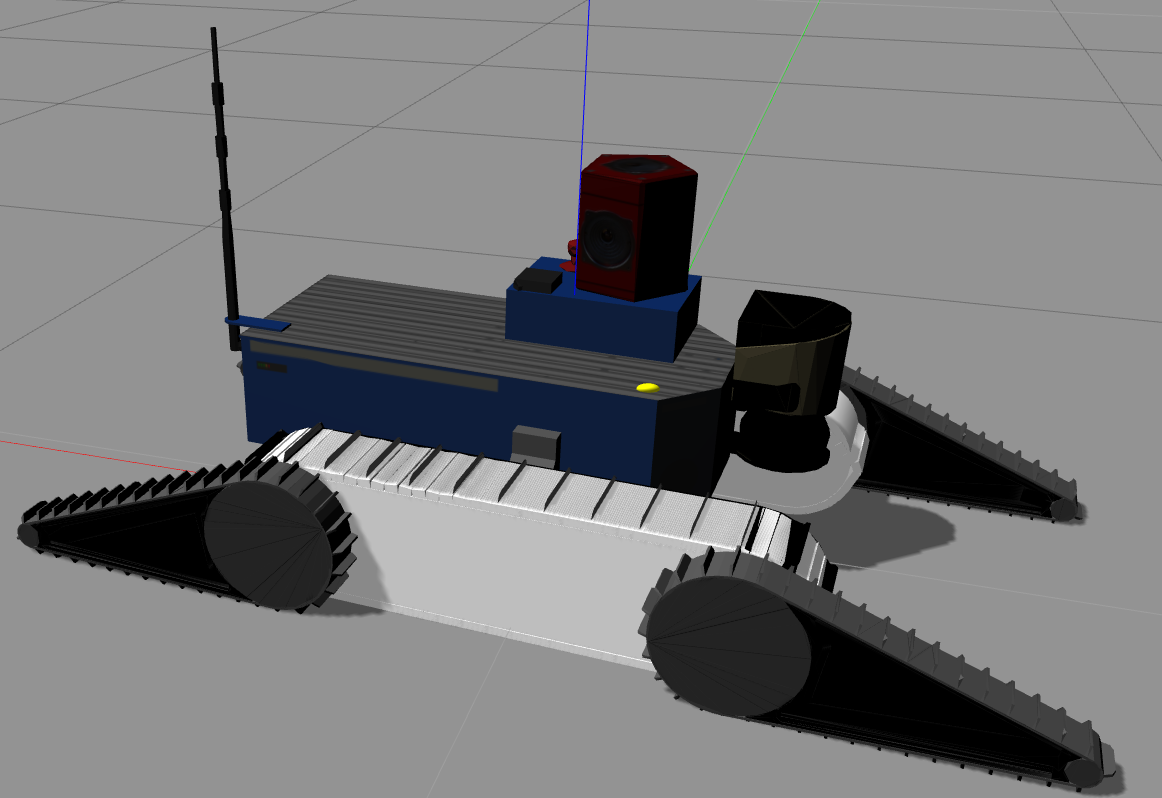}
   	
    \caption{\textbf{Track models.} 
	    \textit{\textbf{Top left}}: A~vehicle with chain-like deformable tracks. This is the model available in model database of the V-REP simulator (courtesy of Qi Wang).
		\textit{\textbf{Top right}}: Non-deformable track model used for the proposed method.
		\textit{\textbf{Bottom left}}: Track approximated by 4~wheels.
		\textit{\textbf{Bottom right}}: Track made of 2~cm plates with grousers.
	}
    \label{fig:track_types}
    
\end{figure}

\section{TYPES OF CATERPILLAR TRACKS}
To clearly specify the type of vehicles this work is focused on, a~short taxonomy of track types follows.

Based on the material the track is made of, the two basic types are \textit{metal} tracks and \textit{rubber} tracks. 
Metal tracks are usually made of many small \textit{track plates} connected together with hinge-like joints. 
Rubber tracks are made of a~continuous steel-reinforced band of rubber.

Another distinctive feature of different track types is the deformability of the outer shape of the track.
The deformable track systems need a set of inner (sometimes also outer) wheels keeping the track approximately in the required shape and providing suspension (see~\autoref{fig:track_types}).
The track can bend in between the wheels, hence the name \textit{deformable tracks}.
Metal tracks are usually deformable, and also deformable rubber tracks exist.

\textit{Non-deformable tracks} have solid guides (infills), which prevent the outer belt shape from  bending and deformation~(see~\autoref{fig:track_types}).
This design is often chosen for rubber tracks, and it is the type this comparison is focused~on.

A~special category---\textit{conveyor belts} and \textit{escalators}---may be added to this taxonomy.
In many design principles they are similar to the tracks for vehicles, but the main difference is they are always fixed to the environment and thus have no dynamics as a~whole.

Independently from the above categories, tracks can be equipped with \textit{grousers}.
These protrusions enlarge the contact surface and help to increase traction in soft materials (depicted in~\autoref{fig:track_types}).

\section{RELATED WORK}

Depending on the purpose of the simulation, either very precise and detailed, or approximate models can be used.
The former ones have been studied extensively in literature, whereas the approximate models, due to their triviality and inaccuracy, have not been examined profoundly despite their frequent use in nowadays robotics.

\subsection{Precise Models}

Simulation of the deformable tracks can be completely set up using existing robotics simulators -- the track consists of a~set of solid track plates connected with hinge joints, several wheels and, possibly, suspension of the wheels.
All these components are available in simulators like Gazebo, Webots or V-REP.
However, this type of simulations is both computationally intensive and very unstable for the high number of constrained dynamic elements~\cite{Kenwright2012}.
Only the V-REP simulator provides a~reliable simulation of this type, and many parameters have to be very finely tuned for it to work.
Sokolov~et~al.~\cite{sokolov2017} tried to implement this method in Gazebo, but the reported results are unsatisfactory.

When the general-purpose simulators fail, specialized simulators were developed to simulate the deformable track dynamics.
Wallin~et~al.~\cite{Wallin2013} compared several formulations of the mechanical joints when applied to metal tracks.
They conclude that each formulation has its advantages and disadvantages and has to be chosen with respect to the specific use-case.

As discussed in the introduction, considerable effort is devoted to simulation of tracks using the Finite Elements Analysis~\cite{Arias,Ma2006}.
But the precision and computational demands are of higher orders than the methods we focus on.

In agriculture and military research, the track-soil interaction is of high interest (mainly due to sinkage of the track plates).
Most of these works seem to only consider planar motion of the vehicle~\cite{Ferretti1999,Janarthanan2012,Rubinstein2004} and mainly concentrate on computing correct sinkage-induced behavior.
Yamakawa and Watanabe~\cite{Yamakawa2004} provide a fully three-dimensional simulation taking into account the track-soil interactions and wheel suspension.

\subsection{Approximate Models}

Common feature of the models described in the previous section is that they properly simulate some effects, but are either very computationally intensive, or neglect some other important effects (they e.g. assume motion on flat ground with small obstacles only).

We are not aware of any approximate model for the deformable track type, because its behavior is highly nonlinear and it essentially requires to model the individual parts of the track separately.
The rest of this section thus concentrates on approximate models for non-deformable tracks.

In some environments, only flat ground is present (e.g. in household robotics or storehouse helper robots).
Then there is effectively only a very small difference between a~tracked robot and a~4-wheel robot with skid-steer control.

In some cases, the tracks can be treated completely passive and the robot motion can be roughly estimated by setting zero friction to the track surface, and pushing the robot with a~virtual force instead of driving the tracks.
This force can be applied via a P(ID) controller, so that the robot achieves the desired velocity and keeps it.
However, the usual effects of friction can not be simulated.
Consequently, the robot can not stand on a~tilted plane without control force (which the real robot can do).

When negotiation of obstacles needs to be accounted for, the 4-wheel approximation would fail because the robot could not support itself on obstacle edges by the middle parts of the tracks.
In this case, the problem is often solved by putting more virtual intersecting wheels inside the track.
This approach has been tested by Sokolov~et~al.~\cite{sokolov2016}, and is available as a~predefined model in V-REP and Webots simulators.
The model still uses the skid-steer wheel control with synchronized wheel velocities on each side.
On one hand, it has problems imitating the skid-steer behavior properly.
On the other hand, the robot is able to overcome some obstacles and can support itself by any part of the track.
But the geometry of such model does not correspond to the real geometry, which is why these models cannot plausibly simulate e.g. climbing up a~staircase.
We have observed in~\autoref{sec:experiments} that this model also gets stuck in some cases where the real robot would continue going.
These models also do not work very well with the standard \textit{friction pyramid} approximation of friction direction -- it is instead needed to use the more precise (and more computationally expensive) \textit{friction cone} model~\cite{Drumwright2010}.

The V-REP simulator offers another method of approximate simulation, which is only suitable for conveyor belts and other static elements.
It bypasses the physics by directly setting linear velocity of the whole conveyor belt mechanism, letting it interact with other bodies, and resetting all forces that acted on it afterwards.
This way, the conveyor belt can exert forces on objects colliding with it, but at the same time, it stays on its place unaffected by any kind of dynamics (because the forces are zeroed-out each simulation step).

\subsection{Skid-steer Motion}
\label{ssec:skid-steer}

The slippage in the skid-steer behavior is an essential part of motion of tracked vehicles.
While it automatically emerges from the precise simulation models as a~result of track tension and other forces acting on the individual parts of the track, a~kinematic model is also available for approximate or kinematics-only simulations.

Mart{\'{i}}nez~et~al.~\cite{Martinez2005} define virtual points called \textit{Instantaneous Centers of Rotation} (ICR) which depend on the desired turning radius and on a~coefficient called \textit{steering efficiency}.
The robot follows a~circular path centered at the ICR and if the steering efficiency is equal to~$1$, the motion is the same as the motion of a geometrically equal differential-drive wheeled vehicle.

Janarthanan~et~al.~\cite{Janarthanan2011} extend this theory for tracked vehicles with road wheels.

\section{Model Based on Contact Surface Motion}

Our novel method exploits the dynamic simulation formulation as \textit{Linear Complementarity Problem} (LCP), which is used in ODE~\cite{ode} and other robotics simulators.
It does not, however, depend on any particular LCP solver implementation.

\subsection{LCP Formulation}

The dynamic simulation problem is an application of Newton's second law:
\begin{eqnarray}
\label{eq:newton}
\mathbf{F} = M a = \frac{d(M \mathbf{\dot{q}})}{dt}
\end{eqnarray}
where $t$ is time, $\mathbf{F}$ is the force acting on the dynamic system, $M$ is the mass and inertia matrix, and $\mathbf{\dot{q}}$ is the linear and angular velocity of the bodies (which is the derivative of the system state~$\mathbf{q}$).
The force $\mathbf{F}$ is split into \textit{external force} $\mathbf{F_e}$ and \textit{constraint force} $\mathbf{F_c}$~\cite{Kenwright2012}, which is a~set of forces generated by joint constraints that keep joint constraints valid in the next time step.

The constraints are written in the form 
\begin{eqnarray}
\label{eq:constraint}
\dot{C}(\mathbf{q}) = J \mathbf{\dot{q}} \geq 0
\end{eqnarray}
where $J$ is the \textit{constraint Jacobian}. 
An observation in~\cite{Kenwright2012} states that the direction of the constraint force is given by~$J$, so it is sufficient to search for the constraint force magnitude $\lambda$ (so that $\mathbf{F_c}~=~J~\lambda$).

In simulation, the derivative is discretized into short time steps $\Delta t$ (usually 1~ms) and the state of the system is integrated step-by-step using Euler's integration~\cite{Kenwright2012}.
The state of the system in the next time step $n+1$ can be expressed as
$$ \mathbf{q_{n+1}} = \mathbf{q_n} + \mathbf{v_{n+1}} \Delta t$$
where the new velocity vector $\mathbf{v_{n+1}}$ (corresponding to $\mathbf{\dot{q}}$ in the continuous setting) is obtained from~\autoref{eq:newton}:
\begin{eqnarray*}
\mathbf{v_{n+1}} & = \mathbf{v_n} + M^{-1} (\mathbf{F_e} + \mathbf{F_c}) \Delta t\\
& = \mathbf{v_n} + M^{-1} (\mathbf{F_e} + J \lambda) \Delta t
\end{eqnarray*}

The unknown constraint force magnitude $\lambda$ is the solution of the following LCP~\cite{Kenwright2012}:
\begin{eqnarray*}
J M^{-1} J^T \lambda \Delta t + J (\mathbf{v_n} + M^{-1} \mathbf{F_e} \Delta t) & \geq 0 \\
\mathrm{given}~\lambda \geq 0,~J (\mathbf{v_n} + M^{-1} \mathbf{F_e} \Delta t) & \geq 0 \\
(J (\mathbf{v_n} + M^{-1} \mathbf{F_e} \Delta t))^T \lambda & = 0
\end{eqnarray*}

\subsection{Contact Constraint Equations}

In each time step, when links $L_1$ and $L_2$ collide, a set of contact points $\{C_i\}_{i=0}^{N}$ is generated at places where the links touch or penetrate each other.
Every contact point is assigned a~\emph{contact joint}, which is a temporary constraint between $L_1$ and $L_2$.
The set of constraints yielded by the contact joint consists of a position constraint (repelling the two links from each other along the contact normal), and a velocity constraint for friction (stopping parallel motion of the two links), which often utilizes the Coulomb friction representation~\cite{Trinkle1997,KSJP:2008}.

Linear velocity of $L_1$ is denoted by~$\mathbf{v_1}$, angular velocity by~$\mathbf{\omega_1}$, and $\mathbf{r_{1i}}$ is the vector from the center of $L_1$ to $C_i$; respective definitions hold for $L_2$.
Further, $\mathbf{t_i}$~denotes the main tangential friction direction (which is perpendicular to the contact normal).

The approximate velocity constraint for Coulomb friction at contact point $C_i$ with friction coefficient~$\mu_i$ is~\cite{Trinkle1997}:
\begin{eqnarray}
\label{eq:friction}
\frac{\partial C_i}{\partial t} = (\mathbf{v_2} + \omega_2 \times \mathbf{r_{2i}} - (\mathbf{v_1} + \omega_1 \times \mathbf{r_{1i}}))  \cdot \mathbf{t_i} = 0 \\
-\mu_i \leq \lambda_i \leq \mu_i
\end{eqnarray}
which can be interpreted as ``stop any motion in direction~$\mathbf{t_i}$''.
The LCP solver tries to find magnitude of the friction force in direction $-\mathbf{t_i}$ (which is bounded by $\mu_i$) that would satisfy this equation.

\subsection{Contact Surface Motion Model}

With the previous definitions, our novel method can be described as a~modification of~\autoref{eq:friction}.
To account for the track velocity $v_t$, \autoref{eq:friction} is adjusted to:
$$\frac{\partial C_i}{\partial t} = v_t$$
which might be interpreted as ``find a force that would keep relative motion of $L_1$ and $L_2$ at velocity $v_t$''.
With this change, the model will move just by applying the modified friction constraints and setting $v_t$.


Nevertheless, this model is not able to correctly simulate grousers.
If the real track has grousers, one way to add a~similar effect to the simulation is to increase the friction coefficient.
This method proved useful in our previous work~\cite{Pecka-IROS-2016} where we heavily utilized the simulator to find a~control policy suitable also for the real robot.

There are more precise models for contacts with friction~\cite{KSJP:2008}, but the practical experiments have shown that even the friction pyramid approximation used in ODE is sufficient for our method to work.

This method can be also easily used for tracks of various shapes. 
The only requirement is to be able to compute the normals of contact points on the tracks.

\subsection{Enabling Skid-Steer Motion}
The last part to be defined is the friction direction $\mathbf{t_i}$. 
If only forward motion is required, it can be simply set to be parallel to the tracks.
However, this setting causes problems when the robot is to turn around using skid-steering motion (since the friction forces are not consistent with the turning maneuver).

\begin{figure}[b]
    \centering
    
    \includegraphics[height=3cm]{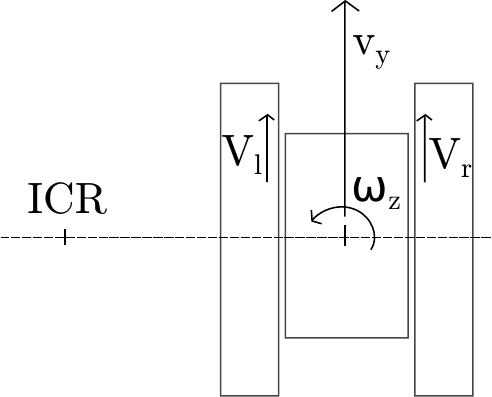}
    \hspace{0.8cm}
    \includegraphics[height=3cm]{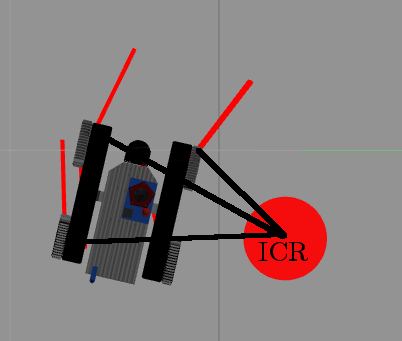}
    
    \caption{\textbf{Instantaneous Center of Rotation.} \textbf{\textit{Left}}: A schematic view of the $ICR$. If the vehicle doesn't slip to the sides, $ICR$ lies always on the depicted horizontal line passing through the centers of the tracks~\cite{Martinez2005}. The distance of $ICR$ from the center depends on forward velocity $v_y$ and angular velocity $\omega_z$ (inverse kinematics), or the speeds of the left and right track $V_l$ and $V_r$ (forward kinematics).
    \textbf{\textit{Right}}: Computed directions of the friction forces $\mathbf{t_i}$ (red lines) for the case where $ICR$ lies in the center of the red disk. The friction forces are perpendicular to the (black) lines connecting the contact points with $ICR$.}
    \label{fig:icr}
\end{figure}

Here we connect the dynamic simulation with the~kinematic model of tracked vehicle motion by Mart\'{i}nez ~et~al. introduced in~\autoref{ssec:skid-steer}.
The whole vehicle is said to be following a circular path centered at $ICR$ (or driving straight if $ICR$ is in infinity).
Thus, we know the desired trajectory of all contact points on the track, and we set each direction $\mathbf{t_i}$ to be tangent to this trajectory, see~\autoref{fig:icr}.


This model has been successfully used in our previous work~\cite{Pecka-IROS-2016}.
Implementation of the proposed method (and some other) has been offered to the Gazebo community~\cite{gazebo-pr}.

\section{COMPARISON OF MODELS}
\label{sec:experiments}

\begin{table*}[t]
    \centering
    \caption{Numerical comparison of the simulation methods.}
    \label{tab:results}
    \renewcommand{\arraystretch}{1.2} 
    \setlength\tabcolsep{3 pt} 
    \begin{tabular}{|l|l|l|l|l|l|l|l|l|l|l|}
\hline
 & \textbf{Metric} & \textbf{csm} (proposed) & \textbf{4wheels} & \textbf{8wheels} & \textbf{no\_friction} & \textbf{plates10} & \textbf{plates2} & \textbf{plates10g} & \textbf{plates2g} \\
\hline
\textbf{Straight} & $d_{t}$ & $\mathbf{0.1\pm0.0}$ & $\mathbf{0.1\pm0.0}$ & $\mathbf{0.1\pm0.0}$ & $0.2\pm0.1$ & $1.6\pm0.0$ & $1.3\pm0.0$ & $1.6\pm0.1$ & $0.5\pm0.0$ \\
\hline
\textbf{Rotate} & $d_\omega$ & $\mathbf{0.1\pm0.0}$ & $0.5\pm0.0$ & $1.6\pm0.0$ & $\mathbf{0.1\pm0.1}$ & $1.4\pm0.6$ & $1.0\pm0.1$ & $2.5\pm0.2$ & $3.1\pm0.0$ \\
\hline
\textbf{Circular} & $\sum d_{t}$ & $157.8\pm5.7$ & $\mathbf{45.5\pm1.7}$ & $116.9\pm6.4$ & $47.4\pm2.2$ & $210.5\pm30.6$ & $189.5\pm4.7$ & $564.9\pm36.3$ & $195.7\pm6.5$ \\
\hline
\textbf{Back\&forth} & $d_{st}$ & $\mathbf{0.0\pm0.0}$ & $\mathbf{0.0\pm0.0}$ & $\mathbf{0.0\pm0.0}$ & $0.1\pm0.0$ & $1.3\pm0.1$ & $0.1\pm0.0$ & $2.6\pm0.2$ & $0.3\pm0.0$ \\
\hline
\textbf{Ramp} & $\sum d_\omega$ & $\mathbf{1.8\pm0.0}$ & $2.2\pm0.1$ & $\mathbf{1.8\pm0.0}$ & $2.0\pm0.0$ & $4.6\pm1.1$ & $2.6\pm0.4$ & $10.5\pm4.1$ & $34.7\pm1.9$ \\
\textbf{} & $\sum d_{t}$ & $8.0\pm2.7$ & $\mathbf{4.6\pm1.3}$ & $5.3\pm0.8$ & $16.8\pm4.3$ & $36.3\pm1.6$ & $61.7\pm0.2$ & $36.9\pm3.5$ & $121.9\pm1.1$ \\
\hline
\textbf{Staircase} & $\sum d_\omega$ & $14.0\pm0.8$ & $10.2\pm0.1$ & $10.7\pm0.3$ & $17.1\pm0.4$ & $36.0\pm31.3$ & $\mathbf{8.4\pm1.8}$ & $25.1\pm11.1$ & $45.5\pm1.8$ \\
\textbf{} & $\sum d_{t}$ & $\mathbf{12.1\pm1.0}$ & $16.3\pm2.3$ & $15.2\pm0.9$ & $13.0\pm3.7$ & $111.5\pm5.9$ & $86.8\pm2.5$ & $14.4\pm7.0$ & $163.0\pm1.6$ \\
\hline
\textbf{Stand on st.} & $d_{st}$ & $\mathbf{0.0\pm0.0}$ & $0.1\pm0.0$ & $0.2\pm0.0$ & $0.2\pm0.0$ & $0.6\pm0.6$ & $0.2\pm0.0$ & $0.2\pm0.2$ & $0.2\pm0.0$ \\
\textbf{} & $d_\omega$ & $\mathbf{0.0\pm0.0}$ & $\mathbf{0.0\pm0.0}$ & $\mathbf{0.0\pm0.0}$ & $0.1\pm0.0$ & $0.4\pm0.4$ & $0.1\pm0.0$ & $0.2\pm0.2$ & $0.1\pm0.0$ \\
\hline
\textbf{Pallet} & $\sum d_\omega$ & $\mathbf{27.6\pm0.9}$ & $27.7\pm12.5$ & $28.3\pm17.7$ & $31.8\pm1.0$ & $164.4\pm151.5$ & $47.6\pm4.7$ & $83.0\pm13.1$ & $64.3\pm2.6$ \\
\textbf{} & $\sum d_{t}$ & $49.9\pm2.5$ & $116.9\pm76.2$ & $157.8\pm46.6$ & $\mathbf{45.2\pm3.4}$ & $508.3\pm146.3$ & $181.1\pm3.8$ & $198.6\pm72.7$ & $78.4\pm1.2$ \\
\hline
\hline
\textbf{CPU time} & time & $38.9 \pm 1.4$ & $47.5 \pm 1.7$ & $82.4 \pm 3.3$ & $\mathbf{33.0 \pm 1.3}$ & $254.9 \pm 4.9$ & $2282.6 \pm 112.7$ & $203.5 \pm 8.3$ & $2241.3 \pm 31.8$ \\
\hline

    \end{tabular}
    \\
\begin{flushleft}
    Numerical results of the conducted experiments.
    Each model-scenario pair was executed 10~times, and the averages and standard deviations of the defined metrics are shown in the table.
    Shorthand $d_t$ means the \textit{distance to target point} metric (units are meters), $d_{st}$ is distance from start.
    Term $d_\omega$ denotes the smallest angular offset from target roll-pitch-yaw orientation (units are radians).
    Terms $\sum d_t$ and $\sum d_\omega$ stand for the \textit{sum of positional errors} or \textit{sum of angular errors} respectively (with units meters and radians).
    \textit{CPU time} (in last row) is not a scenario, but as it is aggregated over all scenarios for each model, we display it as a row of values.
    The duration of all scenarios in simulation time is 110~seconds, so a run-time of 30~seconds means the simulation ran at~$\frac{11}{3}$~real-time speed on the test notebook.
    Best results in each scenario are highlighted in bold for better orientation.
\end{flushleft}
\end{table*}

In this section, a~comparison of methods of modeling non-deformable tracks is presented.

\subsection{Tested Models}

The tested models are described in the following sections (and depicted in~\autoref{fig:track_types}).
Each model is shortly introduced, and an~abbreviation for it is defined, which is used throughout the rest of the text and figures.
All the tested models differ only in representation of the main tracks -- all other properties, such as mass, inertia, shape etc. were the same for all models.

With each of the models, identification of the most realistic set of parameters was done.
The optimized parameters were always \textit{linear} and \textit{angular gain} -- ratios that convert control inputs from simulator to velocity commands for the models.
Other parameters were added only for the models they make sense with, and consist of~\textit{steering efficiency} and friction coefficients in the first and second friction direction.

First, we tried to manually find a~suitable set of parameters and estimated the ranges for each of them.
Then we did 5~iterations of optimization, in each of which we examined 5~samples from a multivariate Gaussian distribution centered on the so far best set of parameters (with covariance derived from the estimated ranges).
Examination of each sample consisted of traversing all defined scenarios with model settings taken from the sample, and summing up the weighted metric values (defined further in this section).
To account for the uncertainty in the simulator, each traversal was tried 3~times and the metric value was averaged over these trials.

\subsubsection{Model based on Contact Surface Motion}
This is the novel model shortened as \textbf{CSM}.

\subsubsection{Wheels instead of tracks}
Model with 4~wheels instead of each track (\textbf{4wheels}) or 8~wheels (\textbf{8wheels}).
All wheels are velocity-controlled using a~skid-steer wheel control mechanism (with wheels on each side synchronized in~velocity).

\subsubsection{Subdivision to plates}
Model with belt subdivided into 10~cm plates (\textbf{plates10}) or 2~cm plates (\textbf{plates2}) interconnected by hinge joints, plus sprocket and idler wheels.
Versions with grousers attached to the track plates are shortened as \textbf{plates10g} and \textbf{plates2g}.
The inner space of the track is filled with a solid box which can collide with the track plates, thus emulating the non-deformability of the track.
Only the sprocket wheel is controlled, using torque control.
This model requires more tuning in the simulator.
To simplify it, the sprocket wheel is represented by a cylinder with infinite friction with the track plates (so that it efficiently transfers force to them without the need to model the teeth and their interaction with the plates).
Further, lateral motion of track plates has to be avoided (otherwise, they would slip off the track very easily).
This would be best done with a planar joint, which is however not available in Gazebo/ODE.
As a~workaround, placing two virtual vertical plates to the sides of each track (that collide only with the track plates) yields a~similar behavior (although it is not ideal).

\subsubsection{No friction}
Model with zero friction between the tracks and ground (\textbf{no\_friction}).
The collision shape of the track is the same as in the \textbf{CSM} model, but the friction of the track is set to zero, and the whole model is force-controlled by applying a virtual force at its center of mass.
The applied force is always perpendicular to the vertical axis of the robot.

\subsubsection{The real robot}
The \textbf{real} robot was also part of the test.
It is the Absolem platform used in Urban Search and Rescue project TRADR~\cite{NIFTi-JAR2014}.
Position of the robot in 6D~space was measured by an~IMU combined with track and laser odometry~\cite{Kubelka2012}.

\subsection{Test Scenarios}

\begin{figure}[h]
    \centering
    
    \includegraphics[width=\columnwidth]{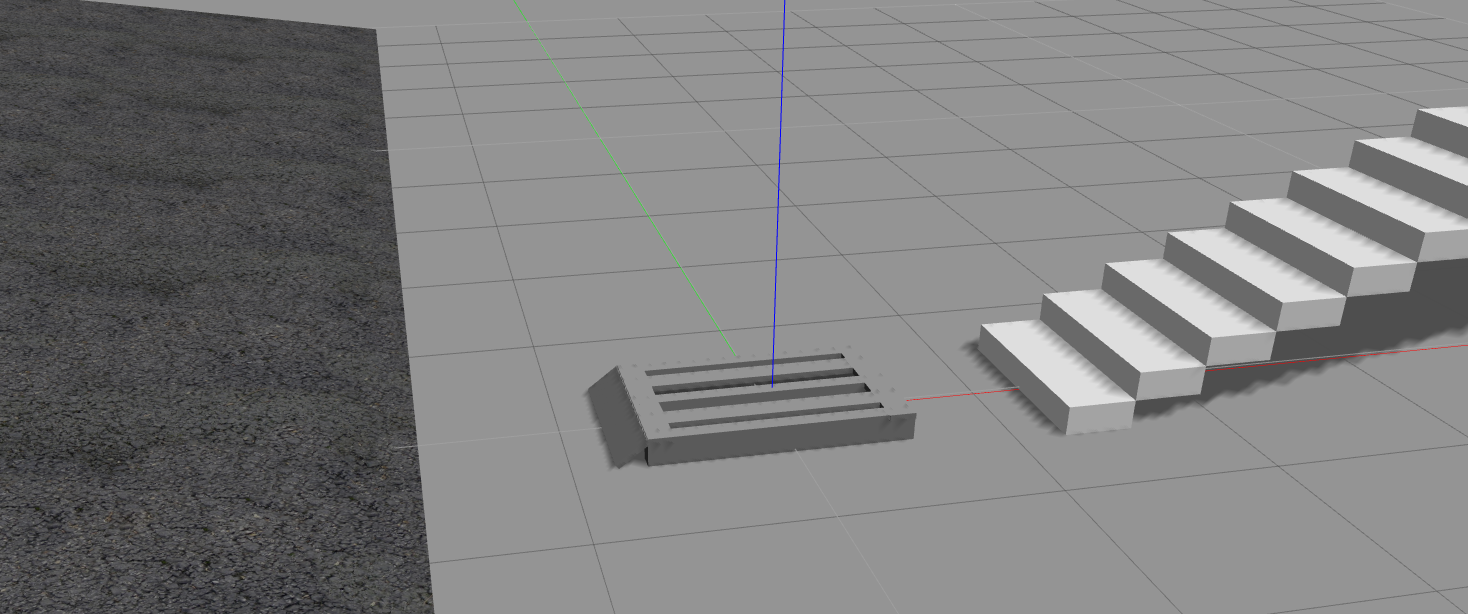}
    
    \caption{\textbf{Obstacles used in test scenarios.} Obstacles that appear in the test scenarios (from the left): ramp, pallet, staircase. Also flat ground was used in scenarios. The models of the obstacles are 1:1~models of the obstacles traversed by the real robot. }
    \label{fig:playground}
\end{figure}

The models were tested in the following scenarios.
Each scenario specifies a different metric showing how successful the model was, and was selected specifically to discover weak points of the models.
All the scenarios start with the robot in rest, no initial speed, forces or torques.
A~view on the obstacles in the scenarios is provided in~\autoref{fig:playground}.

CPU time was measured in all scenarios.
It represents the (real-world) time difference between the start of first scenario execution, and the end of the last scenario execution (so it is summed up over all scenarios for each model).
The simulators were running with high process priority without an upper bound on performance.
The time complexity could be probably lowered for most of the models by adjusting the dynamics engine for the particular case; our measurements show CPU time needed by the implementation in the stock simulator without any code modifications.

Where a~metric refers to the error from real robot trajectory, it means the scenario was traversed with the real robot, and the trajectory was recorded as a~reference.

\subsubsection{Straight drive}

Drive straight on a~building floor using velocity $0.3~m . s^{-1}$ for 10~seconds.
Metric: distance from point $(3.0, 0.0, 0.0)^T$.

\subsubsection{Rotating in place}

Keep the center at one place while rotating at~$0.6~rad . s^{-1}$ for 10~seconds.
Metric: Angular distance from heading $6.0~rad$, metric distance from the starting point.

\subsubsection{Circular path}

Follow a~circular path by driving left track at velocity $0.1~m . s^{-1}$ and right track at velocity $0.3~m . s^{-1}$ for 10~seconds.
Metric: Sum of positional errors (from real robot trajectory) sampled at 10~Hz.

\subsubsection{Ramp}

Drive straight on a~tilted ramp using velocity $0.3~m . s^{-1}$ for 10~seconds.
Metric: Sum of positional errors sampled at 10~Hz, sum of angular errors sampled at 10~Hz.

\subsubsection{Staircase}

Climb down a~staircase using velocity $0.3~m . s^{-1}$ for 10~seconds.
Metric: Sum of positional errors (from real robot trajectory) sampled at 10~Hz, sum of angular errors sampled at 10~Hz.

\subsubsection{Stand on staircase}

Stand on a~staircase with no control commands for 10~seconds.
Metric: Distance from the starting point, angular offset from the starting orientation.

\subsubsection{Pallet}

Climb over a~pallet using velocity $0.1~m . s^{-1}$ for 30~seconds.
Metric: Sum of positional errors (from real robot trajectory) sampled at 10~Hz, sum of angular errors sampled at 10~Hz.

\subsubsection{Back and forth}

Drive using velocity $0.2~m . s^{-1}$ back and forth 10 times, with 2~seconds between every direction switch.
Metric: distance from the starting point.

\subsection{Test results}

\begin{table}[t]
    \centering
    \caption{Summary results}
    \label{tab:results2}
	\renewcommand{\arraystretch}{1.2}
	\setlength\tabcolsep{3.5 pt} 
    \begin{tabular}{|l|c|c|c|c|}
    \hline  & \textbf{CSM} & \textbf{Wheels} & \textbf{Plates} & \textbf{No fric\-ti\-on} \\ 
    \hline \textbf{Computa\-tion speed} & \checkmark  & \checkmark & $\times$ & \checkmark \\ 
    \hline \textbf{Plausibi\-li\-ty on flat sur\-fa\-ces} & \checkmark & \checkmark & \checkmark & \checkmark \\ 
    \hline \textbf{Plausibi\-li\-ty on rough ter\-rain} & \checkmark & $\times$ & \checkmark & $\times$ \\ 
    \hline \textbf{Non-deformable tracks} & \checkmark & \checkmark & \checkmark & \checkmark \\ 
    \hline \textbf{Deformable tracks} & $\times$ & $\times$ & \checkmark & $\times$ \\ 
    \hline \textbf{Grousers} & $\times$ & $\times$ & \checkmark & $\times$ \\ 
    \hline 
    \end{tabular}
    \\
    \begin{flushleft}
    This table presents an overview based on the results of the conducted experiments.
    	Sign ``\checkmark'' means that the model is suitable for/supports the given use-case.
    	Sign ``$\times$'' means that the method is not suitable for/does not support the given use-case.
    \end{flushleft}

\end{table}

Each model was tested 10~times in each scenario, and the values of the metrics were averaged over these tests.

The detailed results are shown in~\autoref{tab:results}.
A~summary extracted from the test results is given in~\autoref{tab:results2}.

From the table, it follows that the track plate models are slower by an order of magnitude or two than the other models.
We have also observed, that the 10~cm plates are too rough approximation of the smoothly curved belt, and the resulting model's motion could be described as ``bumpy''.
Last observation for track plate models is that without grousers, the robot is often not able to climb up the pallet.
That, however, corresponds to the expected real behavior of a~belt without grousers.

The wheeled models are computationally fast and provide good plausibility in most scenarios.
They suffer from unrealistic slippage in the \textit{stand on staircase} scenario, because the friction forces have unrealistic directions.
The \textit{pallet} scenario showed to be a~big problem for these methods---if a~sharp edge (e.g. a~step or pallet edge) touches the track in a~point where neighboring wheels intersect, the model suddenly stops moving as a~result of unrealistic forces and their directions.
We think it is not a~bug in our implementation, since the same behavior was also observed with the wheeled track model available in V-REP simulator (which even uses a~different dynamics engine---Bullet).

The \textit{no\_friction} model provided good results in all tested scenarios, except \textit{stand on staircase}.
That failure is obviously caused by the missing friction between tracks and ground.
It was the fastest tested model.

The proposed \textit{Contact Surface Motion} model was the second fastest tested model.
It provided good results in all tested scenarios except \textit{circular path}.  
Here, the parameter optimization was not able to find a~set of parameters that would provide good performance for both \textit{rotate in place} and \textit{circular path}; with the best set of parameters, the robot was turning too quickly in the \textit{circular path} scenario.
Together with \textit{no\_friction}, only these two models traversed the pallet without problems.

\section{CONCLUSION AND FUTURE WORK}
Simulation of tracked vehicles is a~complicated task even when it is narrowed down only to simulation of non-deformable tracks.
The presented \textit{Contact Surface Motion} model proved to be one of the fastest methods that still provide highly plausible results in most cases.
It is the first computationally-light method allowing the use of precise geometry of the tracks while keeping plausible dynamic behavior.
It can be utilized not only for simulation of tracked vehicles, but also for conveyor belts, treadmills and any other kind of moving planar surfaces.

The proposed set of metrics for comparison of the simulation models showed as a~practical test for discovering weak and strong points of each model.
Once the pull request to Gazebo~\cite{gazebo-pr} is merged, the testing world and obstacles~\cite{dataset} can be utilized by others to compare with their models.

\section*{ACKNOWLEDGMENT}

The research leading to these results has received funding from the European Union under grant agreement FP7-ICT-609763 TRADR; from the Czech Science Foundation under Project GA14-13876S, and by the Grant Agency of the CTU Prague under Project SGS16/161/OHK3/2T/13.

\bibliographystyle{IEEEtran}
\bibliography{IEEEabrv,root}

\end{document}